\documentclass[10pt,twocolumn,letterpaper]{article}

\usepackage{wacv}
\usepackage{times}
\usepackage{epsfig}
\usepackage{graphicx}
\usepackage{amsmath}
\usepackage{algorithmic}
\usepackage{algorithm}
\usepackage{amssymb}
\usepackage{booktabs}
\usepackage[backend=bibtex]{biblatex}

\def\wacvPaperID{****} 

\usepackage[breaklinks=true,bookmarks=false]{hyperref}

\bibliography{main}
\begin{document}

\title{Evaluating Pretrained models for Deployable Lifelong Learning}

\author{Kiran Lekkala, Eshan Bhargava, Yunhao Ge, Laurent Itti\\
\texttt{klekkala@usc.edu}, \texttt{ebhargav@usc.edu}, \texttt{yunhaoge@usc.edu}, \texttt{itti@usc.edu}\\
Thomas Lord Department of Computer Science\\
University of Southern California\\
\footnotesize First two authors contributed equally}

\maketitle
\thispagestyle{empty}

\begin{abstract}
We create a novel benchmark for evaluating a Deployable Lifelong Learning system for Visual Reinforcement Learning (RL) that is pretrained on a curated dataset, and propose a novel Scalable Lifelong Learning system capable of retaining knowledge from the previously learnt RL tasks. Our benchmark measures the efficacy of a deployable Lifelong Learning system that is evaluated on scalability, performance and resource utilization. Our proposed system, once pretrained on the dataset, can be deployed to perform continual learning on unseen tasks. Our proposed method consists of a Few Shot Class Incremental Learning (FSCIL) based task-mapper and an encoder/backbone trained entirely using the pretrain dataset. The policy parameters corresponding to the recognized task are then loaded to perform the task. We show that this system can be scaled to incorporate a large number of tasks due to the small memory footprint and fewer computational resources. We perform experiments on our DeLL (Deployment for Lifelong Learning) benchmark on the Atari games to determine the efficacy of the system.

\end{abstract}


\section{Introduction}
Humans have an innate ability to sequentially learn and perform new tasks without forgetting them, all while leveraging prior knowledge during this process. Continual learning is an imperative skill that needs to be acquired by any intelligent machine. This is especially true in the real-world, where environments keep evolving; thus, agents need to remember previously executed tasks in order to perform these tasks in the future without forgetting. Current continual learning methods use complex memory modules and data augmentations that become difficult to scale and deploy on real-world robotic systems. Furthermore, it's important that models are pretrained offline using large datasets so that when deployed they offer good inductive bias to warm-start the learning process.

\begin{figure}
\centering
    \includegraphics[width=\columnwidth]{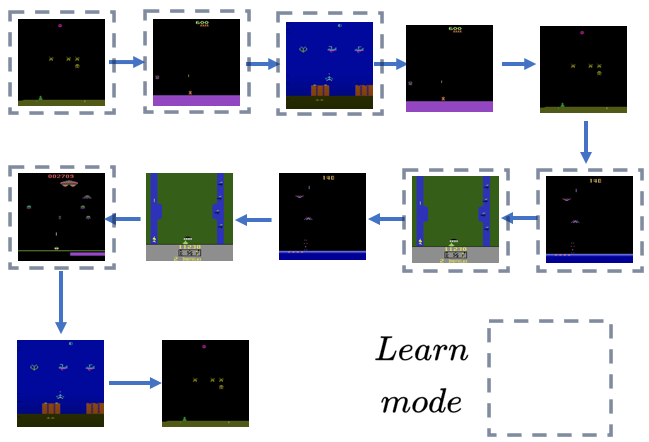}
    \caption{An example run for a typical Lifelong Learning system. The agent is given tasks sequentially with an objective of maximizing performance and minimizing total training time and resource utilization. Training time is counted only during the \textit{Learn mode}.}
    \label{fig:1}
\end{figure}



As an attempt to solve the Lifelong Learning problem for Visual Reinforcement Learning (RL), especially those having more representation complexities than the control, we propose a simple, yet efficient Lifelong Learning system that can be pretrained on a large dataset offline and deployed on a real-world system. The core of our system consists of a meta task-mapper that learns to identify tasks, even when new tasks are given on the fly. Our method's primary novelty lies in the fact that the system is pretrained on a dataset and performs continual learning on a benchmark, with no overlap between both the data distributions.


\begin{figure*}
    \centering
    \includegraphics[width=\textwidth]{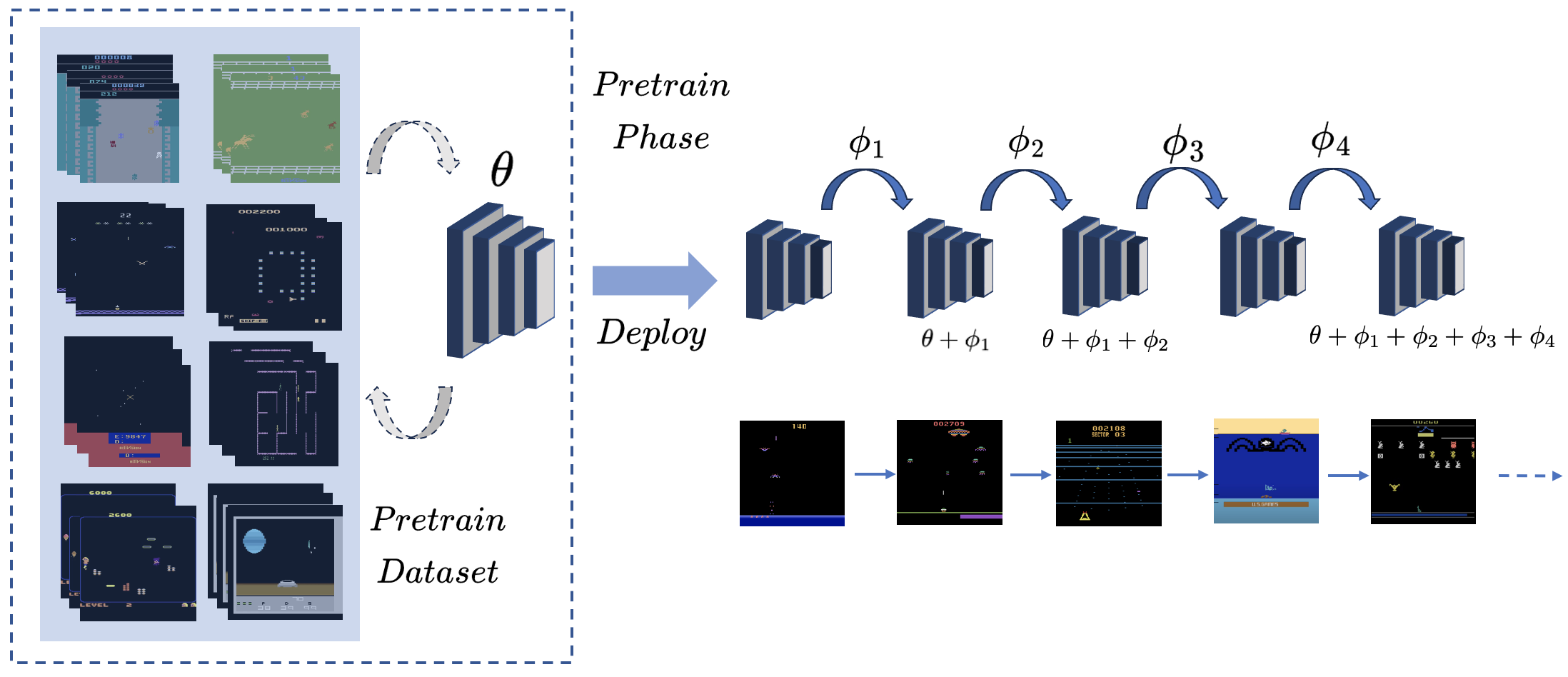}
    \caption{Evaluating \textit{Deployable Lifelong Learning} system involves pretraining a system on a dataset and then deploying it. This process consists of freezing the model parameters and allowing the system to quickly learn and adapt to unseen tasks on the fly. As we can see in the above figure, during deployment, the model sequentially accommodates continual learning on a variable number of tasks by adding a few parameters or datapoints in the data-buffer.}
    \label{fig:second}
\end{figure*}

Lifelong Learning has gained massive popularity in the recent years. \cite{DBLP:journals/ral/LiuXS21} proposes a self-improving Lifelong Learning framework for mobile robot navigation to improve behaviour purely based on its own experience and retain the learnt tasks, but such robots have to retain the experiences of the previous environments. Methods like CRIL \cite{DBLP:conf/iros/GaoGGZ021} apply deep generative replay to alleviate \textit{Catastrophic Forgetting} by generating pseudo-data to train new tasks. Deep \textit{Reinforcement learning (RL)} amidst Lifelong Learning is explored by \cite{DBLP:journals/corr/abs-2202-06843}. Progressive networks \cite{DBLP:journals/corr/RusuRDSKKPH16} start with a single column and a new column for new tasks, although this method is limited by parameters growing faster than the number of tasks. \cite{DBLP:journals/corr/abs-2202-06843} introduces hypernetworks, a meta-model that generates the parameters of a target network that solves the task by using a trainable task embedding vector as an input.

\cite{DBLP:conf/collas/XieF22} follows a similar setting as ours as it uses pretraining and then online learning, however only the policy and the critic are pretrained and data collected is mixed and filtered. Likewise, DARC \cite{DBLP:conf/iclr/EysenbachCALS21} uses domain adaptation and transfer-learning. The model can overcome the difference between the source and target environment, including dynamics, by estimating $\delta_r$ using a pair of binary classifiers. 
 Online adaption or forward transfers are explored in \cite{DBLP:conf/iros/KaushikAM20, DBLP:conf/iclr/Al-ShedivatBBSM18}.  Distillation-based methods \cite{DBLP:journals/corr/RusuCGDKPMKH15, DBLP:journals/corr/abs-2204-05893} are well suited for model/data compression, imitation and multi-task learning. \cite{DBLP:journals/corr/abs-2204-05893} involves model distillation by keeping the policy closer to the behaviour policy that generated the data.

\begin{figure*}
    \centering
    \includegraphics[width=\textwidth]{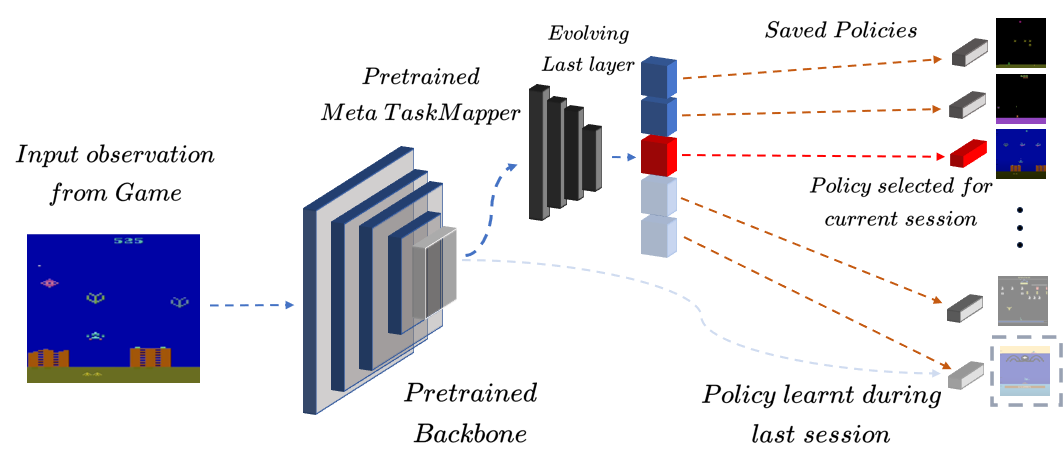}
    \caption{Proposed architecture of our system. Our system contains an encoder and a task-mapper that are pretrained on a large offline dataset. When deployed, our system can identify the previously learned task using the observation from the game. By detecting the task, the appropriate policy can be loaded. On the top, we have the task-mapper, whose last layer is adapted based on the policies the model is currently learning. Arrows and the red modules represent the current policy, selected and loaded, by the task-mapper.}
    \label{fig:second}
\end{figure*}

Offline pretraining is a fast growing field that involves using unlabeled, unorganized data that can be used to learn a pretrained representation model \cite{DBLP:conf/iclr/Al-ShedivatBBSM18}. This model can be used to learn the inductive bias of tasks (like temporal sequencing), relationships of actions with states, and value function estimates corresponding to a state. Currently, the only forward transfer we have in our system is the priors that the backbone/encoder model learns during the offline pretraining, but we are currently working on improving transfer within the games pertaining to the same type.

Existing Lifelong Learning benchmarks evaluate many aspects of the system. \cite{DBLP:conf/collas/PowersXKM022} is one of the first few proposed benchmarks for RL. Compositional Lifelong Learning, like \cite{DBLP:conf/collas/MendezHGE22}, evaluates on the functional aspects of Lifelong Learning. In the OpenAI Atari suite, Gym Retro \cite{DBLP:journals/corr/abs-1804-03720} consists of a large-scale game emulator that has over 1000 games, which could be used to train RL agents. Unlike the above, our benchmark primarily focuses on scalability and resource utilization for deployable Lifelong Learning systems. \textit{The following are the outlined contributions for this paper:}

\begin{enumerate}

\item We collect YouTube videos of Atari games (not in the OpenAI Atari suite) played by human experts and create a dataset that is used by a Lifelong Learning system for pretraining. The model is evaluated on sequential learning of unseen games that are based on OpenAI Gym, and have no overlap with games in the pretrain dataset.
\item To evaluate \textit{Deployable Lifelong Learning} system on performance and resource utilization, we propose a novel benchmark. The code and leaderboard for using the benchmark are made available.

\item Lastly, we propose a novel method that uses the above dataset and benchmark. Our method is based on Few Shot Class Incremental Learning (FSCIL) to learn task differences from the pretrain dataset collected offline and quickly generalize to unseen games.

\end{enumerate}


\section{Dataset and Benchmark}

Before describing our proposed method for Scalable and Deployable Lifelong Learning, we first detail our dataset and benchmark for evaluation. These could be used by researchers to evaluate their models to asses the ability when deployed on real-world systems.

\subsection{Dataset for Pretraining}

To pretrain the system, we collected a dataset using expert-played YouTube videos, such that every video was extracted at 10fps to obtain a sequence of observations. All of these games are different from the games in OpenAI Atari suite. A total of 1,116,275 images with a dimension of $360 \times 480$ were collected as part of the pretrain dataset. All the images were cropped and resized to $84 \times 84$. The list of the games and the format of the dataset can be found here \footnote{\url{https://klekkala.github.io/atari101}}

For each video frame, we also extract associated rewards directly from the frame. Every Atari game consists of the score that is awarded from the start of the game. We use Tesseract OCR engine \cite{10.5555/1288165.1288167} by providing the bounding box of the reward location for each video in order to obtain the reward value for each frame. All the rewards are normalized across both the video and the game by only computing the difference of rewards of the frames as given in the equation below:


\begin{equation}
r^t_{norm} =\left\{
\begin{aligned}
0 & , & r^t - r^{t-1} = 0, \\
1 & , & r^t - r^{t-1} > 0.
\end{aligned}
\right.
\end{equation}

The dataset consists of $101$ folders, each corresponding to a specific game. Each folder consists of a set of Numpy files that consists of a sequence of observations along with the normalized reward value.

Along with the pretrain dataset, we also provide a meta file describing each of the games in the pretrain dataset. The meta file consists of the following attributes
\begin{enumerate}
    \item \textbf{Game Name}: Name of the specific game.
    \item \textbf{Game Type}: Genre of the game. Various Atari games are classified under genres like \textit{Shoot'em up}, \textit{Maze} etc.
    \item \textbf{Input Text}: A brief description of the game's objective. This would enable the agent to understand the game and reuse any previously learnt skills.
    \item \textbf{Minimum Reward}: Minimum reward required by the agent to not switch to learn mode. This reward is computed by evaluating a specific game using a frozen, randomly initialized model.
    \item \textbf{Maximum Reward}: Reward obtained by an end-to-end trained agent. We used a CNN Encoder and the PPO Algorithm for training.
\end{enumerate}

\begin{figure*}
    \centering
    \includegraphics[width=\textwidth]{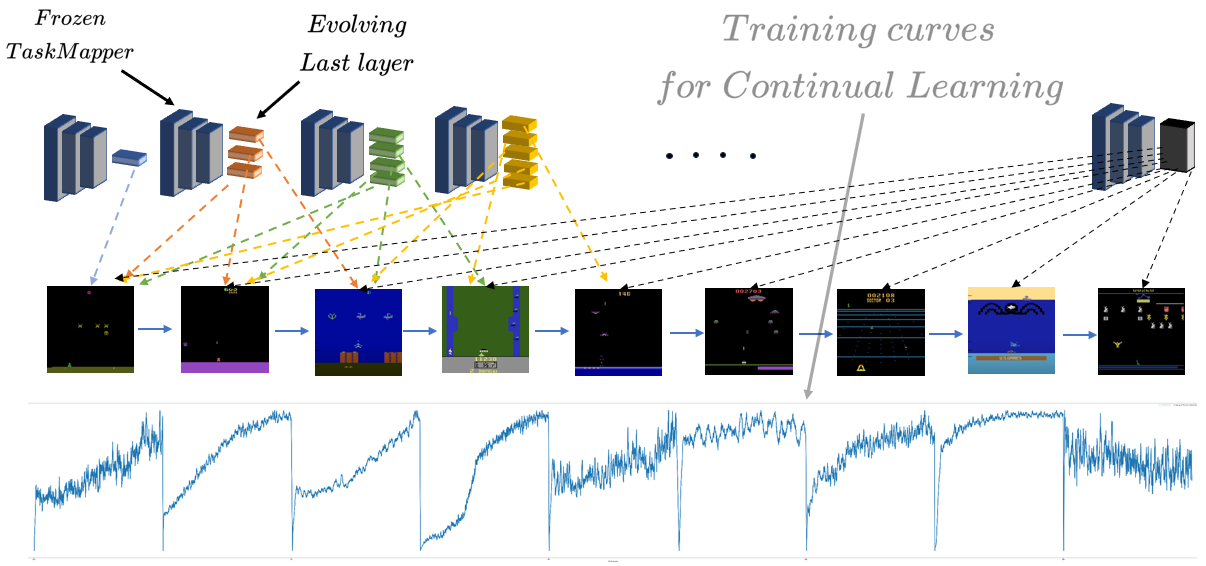}
    \caption{Pictorial overview of our method. With every new game the agent learns, the task-mapper needs to adapt the last layer parameters such that it's able to include the new game in the class prediction. Blue training curves represent the model training on a specific game.}
    \label{fig:second}
\end{figure*}


\subsection{Benchmarking Deployable Lifelong Learning}

Once a model is trained on the above dataset during the pretraining phase, we then evaluate the model's performance on the $DeLL$ benchmark. Note, that we use the term "model" not only for the checkpoint but also as a program for the entire system that takes in an input from the benchmark. The benchmark loads the pretrained model and performs evaluation.

A specific Benchmark $DeLL$ is parameterized by $\alpha$ and $\beta$. $\alpha$ corresponds to the total number of unique games present in the benchmark, and $\beta$ corresponds to the total number of games the agent is given one after the other. Note that for all cases, $\beta > \alpha$. A specific $DeLL$ benchmark consists of $.yaml$ file that has a list of games and the specific game type. Its always assumed that all the game types present in the benchmark are also present in the pretrain dataset, although none of the game themselves are present in the pretrain dataset. For example, the games $DemonAttack$ and $SpaceInvaders$ are part of the benchmark and $Xevious$ and $Galaga$ are part of the pretrain dataset, although all of them fall under \textit{Shoot up} games.

\textit{We urge the reader to take note of some terminology that is beneficial for understanding the benchmark}. Firstly, we use the term \textit{task} and \textit{game} interchangeably since we are currently concerned with Atari games. A specific benchmark $DeLL(\alpha, \beta)$ consists of sequentially evaluating the model on $\beta$ games, which we call a $run$. In a run of $\beta$ games, there are $\alpha$ unique games. A model learns or performs inference on any game in the $\beta$ games during a \textit{session}. A high-level overview of a run is presented in Figure \ref{fig:1}.


Any method/model that is evaluated on a specific benchmark yields 4 different metrics. The following metrics are employed to evaluate the Lifelong Learning system that corresponds to a specific benchmark $DeLL(\alpha, \beta)$. $\alpha$ and $\beta$ correspond to the number of unique games and the total number of games the agent is given one after the other, respectively.

\begin{enumerate}
    \item \textbf{Model Size (MS)} (In \textit{MB}): The size of the model when deployed.
    \item \textbf{Model Inference time (MI)} (In \textit{ms}): Mean inference time of the model on all the $\beta$ games in a run.
    \item \textbf{Learn Switches (LS)}: Number of times the agent switches to learn mode. If the agent obtains a score lower than the minimum reward, then the agent switches to learn mode. 
    \item \textbf{Model Growth (MG)}: Every time the model switches to learn mode, and thereby learns a task, there is an increase in buffer size, model size or both. This value estimates the average percentage increase (in \textit{MB}) of the model after every learn switch.
    \item \textbf{Buffer Size (BS)}: A metric (in \textit{KB}) the model uses to avoid \textit{Catastrophic Forgetting}. This could be any form of data, including images, embeddings, rewards and actions.
    \item \textbf{Buffer Growth (BG)} The average percentage increase of the data buffer increase after every learn switch.
    \item \textbf{Mean Avg Reward (MAR)} Measured during evaluation of a game. This is an array metric, where the length of the array corresponds to the total number of unique games ($\alpha$) in a run. This excludes the sessions that the agent used for learning the game.
    \item \textbf{Total Normalized Mean Reward (TNMR)}: We normalize all the values in the MAR array using the minimum and the maximum reward of the corresponding game, and then compute the mean of the array to get TNMR.
\end{enumerate}

\begin{figure}
\centering
    \includegraphics[width=\columnwidth]{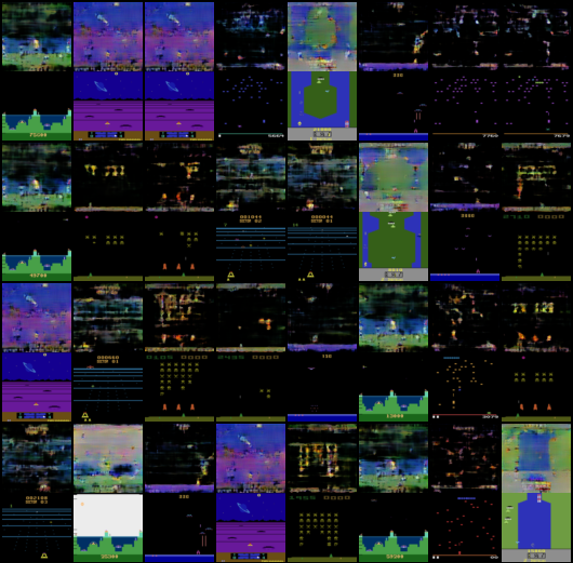}
    \caption{Reconstructions obtained from the trained VAE model on the test-dataset. \textit{Note that the model has not been trained on any of the above games}. Odd rows correspond to the reconstructions of the subsequent even rows.}
    \label{fig:5}
\end{figure}

\section{Proposed Method}
 We propose a Lifelong Learning system designed to identify its originating task using minimal resources, making it well-suited for real-time systems. One of the most important features of our method is that, unlike other methods, it scales well when the number of tasks increases. The most significant novelty of our method is that the entire system is pretrained on a dataset that is different from the dataset used in deployment.

\subsection{Pretrained Encoder}
 We use a pretrained encoder to extract the relevant features from the observations required for task identification and downstream policy execution. During the deployment phase, the pretrained encoder is frozen and used to infer the embeddings, which are then utilized by the task-mapper. Currently, we use a VAE-based encoder that is trained on the pretrain dataset. The reconstructions obtained using the encoder model is presented in Figure \ref{fig:5}.

\subsection{Meta Task-mapper}
One of the major challenges in Lifelong Learning, apart from reusing prior knowledge, is to continually adapt and remember previous tasks. We tackle \textit{Catastrophic Forgetting} by transforming \textit{Continual Learning} into a task identification problem. By predicting the appropriate Task ID during classification, we can load the appropriate policy that was previously learnt, and perform the task. Since the policies are a $1$ layer neural network, the number of tasks can be scaled easily.




We utilize a meta task-mapper that is also trained offline, along with the backbone, to recognize task differences. Given a few observations, the task-mapper learns to identify which of the previously learnt tasks the current task falls into. The task-mapper, denoted by, $g_{\phi}$ parameterized by $\phi$ is trained on a large pretrain dataset. Since the task-mapper has already recognized the differences in the tasks during the pretraining phase, it only needs to adapt to the new tasks using a few-shot learning setting.


 In many real-world instances, the agent needs to keep track of a diverse number of tasks that may keep increasing. In this case, the task-mapper must also accommodate the newer predictions. To allow this, we apply CEC-FSCIL \cite{DBLP:conf/cvpr/ZhangSLZPX21}, which was originally proposed to solve class-incremental continual learning. This method uses a trained graph neural network to learn the correspondences and relations of the classes. During the training phase, the method uses pseudo-incremental training by simulating sequences of different classes in the pretrain dataset. This would mimic how each class would be included in every session during test time. Furthermore the graph model allows a trained task-mapper to be extended to indefinite number of classes by aggregating a list of last-layer parameters corresponding to each class. 
 In the below equation, $N$ corresponds to the $N$-way classification:

\begin{equation}
    \mathbf{W}_{last} = \{w_0, w_1, w_2, w_3, ... , w_N\}
\end{equation}


Once the task-mapper receives the data embeddings for the new sessions, the learnt classifiers in the current session and previous sessions are fed to the graph model for adaptation. The adaptation is done using the support data for $N$ way $K$ shot classification. \textit{At any given point during deployment, the data buffer would consist of $N*K$ datapoints}, where $N$ is the number of learnt tasks (tasks that made the model switch to learn mode). Finally, the updated classifiers can be used for evaluation. As a baseline comparison, we also use a Meta learning \cite{DBLP:conf/icml/FinnAL17} based task-mapper, which unlike our FSCIL based task-mapper has no plasticity. Nonetheless, it can still perform the task-mapping for unseen data during deployment. In which case, for every $N$, there needs to be a different task-mapper. Although this is a naive approach, compared to our task-mapper, we show that even at the expense of more parameters, the CEC-FSCIL based task-mapper outperforms Meta learning based task-mapper for larger values of $N$.

\begin{algorithm}
\caption{System Pre-training}
\begin{algorithmic}[1]
\REQUIRE Offline dataset $\mathcal{D}_{Train}$ containing only sequence of observations $\{o^{i}_1, o^{i}_2, .. , o^{i}_T\}_{i=1}^{M}$
\STATE Train a ResNet VAE using $\mathcal{D}_{Train}$
\STATE Freeze and obtain Encoder $f_{\theta}$ from ResNet VAE
\STATE Initialize Task-mapper $g_{\phi}$
\WHILE{training not done}
\STATE Select train tasks $\tau_i \sim p_{Train}$
\STATE Obtain (Image, class) pairs ${(o_1, c_1), (o_2, c_2), .. (o_K, c_K)}$ corresponding to $\tau_i$
\STATE Estimate and apply gradients using Meta learning or FSCIL loss on $g_{\phi}$
\ENDWHILE
\end{algorithmic}
\end{algorithm}
\begin{algorithm}
\caption{System Evaluation}
\begin{algorithmic}[1]
\REQUIRE Pretrained Backbone $f_\theta$
\REQUIRE Pretrained task-mapper $g_\phi$
\REQUIRE  List of learnt policies $\mathcal{P}$ of size $N$.
\REQUIRE Initialize Task-mapper output to $N$
\REQUIRE Initialize buffer data $\mathcal{D}_{buf}$ containing $N * K$ data-points

\WHILE{not done}
\STATE Request task $\tau$ for evaluation
\IF{mode = \textit{TRAIN}}
\STATE Obtain data $O$ and trained policy $p$ through $O, p \leftarrow$ \textit{RL-Procedure}($\tau$)
\STATE \textit{Selective-Sample} Offline data $O$ into $\mathcal{\hat{O}}$
\STATE Update list of learnt policies $\mathcal{P} \leftarrow \mathcal{P} \cup p$
\STATE Update buffer data $\mathcal{D}_{buf} \leftarrow \mathcal{D}_{buf} \cup \mathcal{\hat{O}}$
\STATE Update output size for the task-mapper $g_{\phi}$ to $N + 1$
\STATE Perform $N$ way, $K$ shot adaptation on $g_{\phi}$
\ELSE
\STATE Obtain test observation(s) $\mathbf{o} \sim \tau$
\STATE Infer Task ID $i$ using the task-mapper $g_{\phi} (o)$
\STATE Load Policy $i$ from $\mathcal{P}$ and perform task $\tau$
\ENDIF
\ENDWHILE
\end{algorithmic}
\end{algorithm}

 Once the backbone and the task-mapper are pretrained on the offline data, using the procedure mentioned in Algorithm 1, they are then deployed on a real-world system and the last layer parameters of the task-mapper are updated using $K$ shot adaptation through the support set. In order to obtain a maximum performance gain using the support set, it's important that we are selective about these $K$ shot support samples from the copious amount of data generated during the RL procedure for learning the new task. Furthermore, by eliminating all the redundant data, we can minimize the computational overhead and accommodate more tasks. Currently, we choose $K$ random samples in a task and store them in a buffer that gets propagated with new sessions, although more optimal methods for selecting the support set may exist. We will be exploring this in our future work. We also show the evaluation/deployment of the system in Algorithm 2.

\begin{table*}[!htb]
    \caption{On the left we see the accuracy of a Meta learning based task-mapper. On the right, we see the accuracy of a CEC-FSCIL based task-mapper. The disadvantage of using a Meta learning based task-mapper is the need for separate last layer for each of the row, whereas in the CEC-FSCIL based task-mapper a single pretrained task-mapper can be used for sequential adaptation.}
    \begin{minipage}{.5\linewidth}
      \label{tab1}
      \centering
\begin{tabular}{|c|c|c|c|c|}

\hline
    & 1-shot & 2-shot & 5-shot & 10-shot \\
\toprule
\hline
\textbf{5-way} & 0.80 & 0.84 &  0.93 & 0.94 \\

\hline
\textbf{10-way} & 0.65 & 0.77 & 0.84  & 0.88  \\

\hline

\textbf{20-way} & 0.55 & 0.62 & 0.75 &  0.80 \\
\hline

 \textbf{30-way} & 0.42  & 0.51  &  0.56& 0.57  \\
\hline

\end{tabular}
    \end{minipage}%
    \begin{minipage}{.5\linewidth}
      \centering
\begin{tabular}{|c|c|c|c|c|c|}

\hline

    & 1-shot & 2-shot & 5-shot & 10-shot & 15-shot \\
\toprule
\hline
\textbf{1-way} & 0.76 & 0.77 & 0.77 & 0.78 & \textbf{0.79}\\

\hline
\textbf{2-way}& 0.74 & 0.78 & 0.78 & \textbf{0.79} & 0.78 \\

\hline

\textbf{3-way} & 0.76 & \textbf{0.78} & 0.77 & 0.77 &  0.77 \\
\hline

 \textbf{4-way} & 0.77 & 0.77 & 0.77 & 0.77 & \textbf{0.78} \\
 \hline
 \textbf{5-way} & 0.78 & 0.76 &  \textbf{0.79} & 0.79 & 0.78 \\

\hline
\textbf{6-way} & \textbf{0.79} & 0.77 & 0.79 & 0.79 & 0.78 \\

\hline

\textbf{7-way} & 0.79 & 0.77 & 0.79 & 0.77 & \textbf{0.81} \\
\hline

 \textbf{8-way}&  0.76 & 0.78 & \textbf{0.80} & 0.78 & 0.78  \\
 \hline
 \textbf{9-way} & 0.78 & \textbf{0.79} & 0.77 & 0.78 & 0.79  \\

\hline
\textbf{10-way} & 0.77 & 0.79 & 0.80 & \textbf{0.81} & 0.77 \\

\hline

\textbf{11-way} & 0.74  & 0.80 & 0.78 & \textbf{0.81} & 0.79 \\
\hline

 \textbf{12-way} & 0.77 & 0.78 & 0.78 & \textbf{0.80} & 0.80 \\
 \hline
 \textbf{13-way} & 0.76 & 0.77 & 0.77 & 0.78 & \textbf{0.79} \\

\hline
\textbf{14-way} & 0.78 & 0.78 & 0.78 & 0.78 & \textbf{0.79} \\

\hline

\textbf{15-way} & 0.81 & 0.78 & \textbf{0.82} & 0.82 & 0.78  \\
\hline

 \textbf{16-way} & \textbf{0.80} & 0.79 & 0.79 & 0.77 & 0.77 \\
 \hline
 \textbf{17-way} & \textbf{0.81}  & 0.80  & 0.81 & 0.78 & 0.78 \\

\hline
\textbf{18-way} & 0.75 & 0.76 & \textbf{0.81} & 0.78 &  x\\

\hline

\textbf{19-way} & 0.77 & \textbf{0.82} & 0.77 & 0.78 &  x \\
\hline

 \textbf{20-way}& 0.77 & 0.80 & 0.81 & \textbf{0.82} & x \\

\hline
\end{tabular}
    \end{minipage} 
\end{table*}



\begin{table*}
\caption{Performance of the proposed system when evaluated on the games sequentially. Note that the reward values are formatted with the respective \textbf{mean/median} value. Random and Trained correspond to the random or pretrained encoders that are frozen. Net-mean total reward corresponds to the average value of the reward after multiplying the CEC accuracy to the Trained-agent reward value. This value estimates the performance of our system during sequential evaluation of the games. Note that for all the sequentially executed tasks, the CEC based task-mapper performs better than the baseline (Meta learning based task-mapper).}
\centering
\begin{tabular}{|c|c|c|c|c|c|c|}

\hline
  \textbf{Name} & \textbf{Random} & \textbf{ML} & \textbf{CIFL} & \textbf{Trained} & \textbf{Net-mean} \\
\toprule
\hline

AirRaid-v4 & 605.0/605.0 & 1.0 & 1.0 & 750.0/712.5 & 750.0 \\

\hline

Assault-v4 & 272.0/268.0 & 0.8 & 0.76 & 300.3/315.0 & 228.2 \\

\hline

BeamRider-v4 & 420.0/425.0 & 0.8 & 0.74 & 440.0/440.0 & 325.6  \\

\hline

Carnival-v4 & 2123.0/2242.0 & 0.8 & 0.76 & 2639.0/2810.0 & 2005.64  \\

\hline

DemonAttack-v4 & 240.0/255.0 & 0.8 & 0.77 & 276.0/257.5 & 212.5 \\

\hline

NameThisGame-v4 & 3923.0/4310.0 & 0.65 & 0.78 & 4052.0/4230.0 & 3160.56  \\

\hline

Pooyan-v4 & 1020.0/1120.0 & 0.65 & 0.79 & 1106.5/997.5 & 874.13  \\

\hline

Gopher-v4 & 732.1/733.0 & 0.65 & 0.76 & 746.0/620.0 & 566.96  \\

\hline

Riverraid-v4 & 2723.0/2740.0 & 0.65 & 0.78 & 2886.0/2815.0 & 2251.08 \\

\hline

Solaris-v4 & 1001.0/1101.0 & 0.65 & 0.77 & 1094.0/840.0 & 842.38  \\

\hline

SpaceInvaders-v4 & 401.0/435.0 & 0.55 & 0.74 & 427.0/385.0 & 315.98 \\
\hline
\end{tabular}
\label{tab2}
\end{table*}

\subsection{Policy}

To perform a specific task, we employ a 1-Layer policy that receives the feature embedding for action. Using a \textit{Float16} quantized format, we store each policy in under 1.5 MB, tagged with its class-id from the task-mapper.

\section{Evaluation and Results}

We use a ResNet-based architecture for the VAE Encoder. The encoder and the decoder block are built for a $84 \times 84$ image and result in a latent vector of size of $512$. The entire network was end-to-end trained for 100 epochs which took about 27 hours on a NVIDIA V100 GPU. The task-mapper consists of a Feed forward Neural Network that has a variable last layer based on the number of learnt tasks. The last layer weights are the only parameters that are continuously updated as the agent keeps learning unseen tasks, and the remaining parameters stay frozen.


\subsection{Evaluation of the system}
We perform experiments by evaluating our system on a set of 11 Atari games from OpenAI Gym. The list of games and their corresponding results are outlined in Table \ref{tab2}. Note that for all these evaluations, the encoder was trained on the pretrain dataset and frozen, and only the policy is trained on the respective game. We also provide the results obtained using our benchmark on our proposed Lifelong Learning system and 3 other baselines (Random encoder, E2E (End to end trained) and Meta learning based task-mapper) in Table \ref{tab3}.

\subsection{Ablation experiments for task-mapper}
The results obtained from the CEC-FSCIL task-mapper are presented in Table \ref{tab1}, alongside a baseline comparison from the Meta learning-based task-mapper. From these tables, we can see that the CEC-FSCIL task-mapper results stay consistent throughout different $N$ way, whereas for the Meta learning based task-mapper performs well with a low $N$ but the performance significantly drops with increasing $N$.

\begin{table}
\caption{Performance of each baseline on our benchmark with $\alpha=5$ and $\beta=10$ parameters. Please refer to Page-5 for the definitions of the metrics. Note that Random and E2E backbones don't have any buffer.}
\label{tab3}
\centering
\begin{tabular}{|c|c|c|c|c|c|}

\hline

    & MS & MG & BS & BG & TNMR \\
\toprule

\hline
\textbf{Random} & \textbf{235} & 0 & 0  & 0 & 0.41  \\

\hline

\textbf{E2E} & 236 & 0 & 0 &  0 & 0.48 \\
\hline

 \textbf{ML} & 257  & 0.51  &  1.2 & 5.6 & 0.67  \\
 \hline

 \textbf{CEC} & 242  & \textbf{0.13}  &  1.2 & 5.7 & \textbf{0.71}  \\
\hline

\end{tabular}

\end{table}



\section{Discussion and Conclusion}

We propose a dataset and benchmark that allows for evaluating deployable Lifelong Learning systems. The dataset consists of sequences of games and rewards obtained from human expert played YouTube videos. A model, when pretrained on the offline dataset, is used as a warm-start to adapt to unseen tasks. Furthermore, we propose a simple, yet scalable framework for Lifelong Learning that involves a \textit{task-mapper} and a backbone that are both pretrained using an offline dataset. The task-mapper evolves with each new task and learns to identify a new task based on previously seen tasks. The entire system is evaluated on a suite of test tasks. Although our method is simple, it scales well, even with a large number of tasks. 

In this paper, we use a representation bottleneck that learns embeddings of the observations solely based on appearance. Even when there are differences in appearances, if the skills learnt are similar, the games can still be played using a single set of parameters. For example, even though \textit{Phoenix} and \textit{AirRaid} have different appearances, they share the same action space and are both \textit{Shoot Up} games. We are currently working on incorporating representations that are skill/dynamic-aware, as opposed to those based solely on appearance, like VAEs. This would not only identify previously learnt tasks, but also help reuse existing policy parameters to play a new game sharing existing skills.


\printbibliography

\end{document}